# A Taxonomy of Errors in *English as she is spoke*:
# Toward an AI-Based Method of Error Analysis for EFL Writing Instruction

Damian Heywood, Joseph Carrier, and Kyu-Hong Hwang

DongA University

Abstract: This study describes the development of an AI-assisted error analysis system designed to identify, categorize, and correct writing errors in English. Utilizing Large Language Models (LLMs) like Claude 3.5 Sonnet and DeepSeek R1, the system employs a detailed taxonomy grounded in linguistic theories from Corder (1967), Richards (1971), and James (1998). Errors are classified at both word and sentence levels, covering spelling, grammar, and punctuation. Implemented through Python-coded API calls, the system provides granular feedback beyond traditional rubric-based assessments. Initial testing on isolated errors refined the taxonomy, addressing challenges like overlapping categories. Final testing used *English as she is spoke* by José da Fonseca (1855), a text rich with authentic linguistic errors, to evaluate the system's capacity for handling complex, multi-layered analysis. The AI successfully identified diverse error types but showed limitations in contextual understanding and occasionally generated new error categories when encountering uncoded errors. This research demonstrates AI's potential to transform EFL instruction by automating detailed error analysis and feedback. While promising, further development is needed to improve contextual accuracy and expand the taxonomy to stylistic and discourse-level errors.

1. Introduction

1.1. Background

Recent developments in artificial intelligence (AI), particularly Large Language Models (LLMs), have shown promise in automating previously unavailable aspects of student writing assessment and providing detailed, individuated feedback. Our previous research demonstrated that AI systems can reliably assess student writing using standardized rubrics, achieving consistency rates of over 99% over five iterations (Heywood & Carrier, 2024). However, while these systems excel at providing holistic assessment using broad categories, their potential to provide detailed, granular feedback about specific writing errors has not yet been fully explored .

This study builds upon our earlier work by developing and testing a sophisticated error classification system that can identify, categorize, and describe writing errors at both the word and sentence levels. The system employs a detailed taxonomy of errors based on established linguistic theory in the area of error classification (Corder, 1967, 1975, 1981; Richards, 1971, 1974; James, 1998). The AI analysis is implemented through carefully designed API calls to Claude 3.5 Sonnet in Python. With this enhanced error classification system, the present study analyzes an error ridden dialogue from an infamous text, *English as she is spoke* (Fonseca et al., 2004). We also provide the results of a review of the AI analysis by a human panel of experts.

1.2. Research objectives

This study seeks to answer the following question: Can an AI system reliably identify and classify specific types of writing errors in a given text using a comprehensive taxonomy of linguistic errors? From this perspective, the study explores whether AI systems can move beyond simple error detection to achieve a deep level of linguistic analysis. A reliable error classification system must distinguish between errors at multiple linguistic levels, from basic orthographic mistakes to complex issues of syntax and discourse. Furthermore, the system must consistently apply a standardized taxonomy that aligns with established theoretical frameworks in second language acquisition.

This level of analysis has significant pedagogical implications, as proper error classification will help instructors understand the developmental stage of language learners and design appropriate interventions. It will also help language learners, as they will have access to extremely specific feedback about their writing at a variety of levels. The question also addresses the technical feasibility of implementing such sophisticated linguistic analysis through current AI technologies, particularly whether currently available Large Language Models (LLMs) can be effectively prompted to perform detailed grammatical analysis with the consistency and accuracy required. If so, this would represent a significant advancement in Ai-based computer-assisted language learning (iCALL), potentially transforming how we approach error correction and feedback in EFL writing instruction.

The hypothesis for the study are as follows:

H1: An AI system can be designed that consistently identifies and classifies the errors in a given text according to an established taxonomy.

H2: The AI system's error classification of a given text is confirmed as accurate by a panel of expert linguists upon review.

1.3. Need for research

While automated writing assessment tools have become increasingly sophisticated, their ability to provide detailed, accurate error analysis remains understudied. A recent comprehensive systematic review of studies concerning Ai-assisted automated writing feedback (AWF) showed that existing research has focused primarily on broad evaluative measures (for example, using banded rubrics) rather than specific error identification and classification (Shi & Aryadoust, 2024). This study addresses these gaps by developing and testing a comprehensive error classification system.

This research has only now become possible due to several advances in the quality and availability of AI technologies. LLM capabilities have rapidly evolved in recent years, particularly in the processing and understanding of natural language, enabling LLMs to better understand context, nuances, and subtleties in language. This progress has been facilitated by improvements in model architectures, training methodologies, and the availability of extensive datasets (Ren, 2024). Powerful models have also become available for use for researchers using complex code-based processes through Application Program Interfaces (APIs) that allow LLMs to be trained on specific sets of corpi and given complex instructions (Wu et al., 2024). This has created new opportunities for developing more sophisticated automated feedback systems.

The implications of this research extend beyond mere technological advancement. If successful, this approach could dramatically increase the quantity and quality of feedback available to students while reducing the time burden on instructors. Furthermore, by providing consistent, detailed error analysis, such a system could help students better understand and correct their writing mistakes, ultimately leading to more effective language acquisition through individuated, needs-specific instruction. The tool will also be useful for linguistics research in general, and points to the potential of coded systems using API calls to perform complex linguistic analysis not previously feasible at this scale.

2. Literature review

2.1. Error analysis in EFL writing: Theory and development

The field of second language acquisition underwent significant theoretical changes during the 1960s and 1970s. The traditional Contrastive Analysis (CA) approach, which attempted to predict learning difficulties by comparing languages, was challenged by the emergence of Error Analysis (EA). While CA attributed most learner errors to first language interference, EA proposed a more sophisticated understanding of the role of errors in language learning.

Corder's 1967 publication "The Significance of Learners' Errors" marked a turning point in how researchers and teachers viewed student errors. Instead of treating errors as evidence of learning failure, Corder argued that they demonstrated active engagement with the target language. He made an important distinction between systematic errors, which reveal the learner's current understanding of the language system, and simple mistakes, which represent temporary lapses in performance. This framework helped establish the concept of interlanguage - a developing linguistic system that changes as learners gain proficiency.

Richards (1971) built on this foundation in "A Non-Contrastive Approach to Error Analysis" by identifying multiple error sources beyond first language interference. His work demonstrated that many errors arise from the inherent complexity of the target language (intralingual errors) or from learners testing hypotheses about language rules (developmental errors). This research established that native language interference, while important, was just one of many factors contributing to language learner errors.

2.2. Major taxonomies and classification systems

Since that time, several major taxonomies have emerged for classifying language learner errors. Burt and Kiparsky (1974) introduced the distinction between global and local errors. Global errors impede communication by affecting overall sentence organization, while local errors affect single elements without disrupting meaning. This classification helped teachers prioritize error correction by focusing on errors that most significantly impact comprehension.

Richards' (1971) taxonomy remains influential in error analysis. He categorized errors into interlingual errors (from first language interference), intralingual errors (from misunderstanding target language rules), and developmental errors (from hypothesis testing). Dulay, Burt, and Krashen (1982) later expanded this framework by adding a surface structure taxonomy that classified errors as omissions, additions, misformations, or misordering.

Expanding on his earlier work, Corder (1981) proposed a systematic approach to error classification based on linguistic levels: phonological, graphological, grammatical, and lexico-semantic. This system proved particularly useful for analyzing written work as it aligned with traditional areas of language instruction. James (1998) further refined these categories and added the dimension of pragmatic errors, acknowledging the importance of contextual appropriateness in language use.

This study employed a detailed error taxonomy based primarily on Corder's (1981) linguistic level classification system, but substantially modified to focus specifically on written production in EFL contexts using a framework adapted from Seddik (2023).

2.3. Pedagogical implications and limitations

The implementation of systematic error analysis in EFL writing instruction offers significant pedagogical potential while presenting several practical challenges. In terms of applications, detailed error taxonomies enable teachers to identify patterns in student writing that can inform targeted instruction. By categorizing errors systematically, instructors can develop focused remediation strategies and create materials that address specific areas of difficulty (Karim et al., 2018). This approach proves particularly valuable in EFL contexts where students often share common first language influences on their writing.

However, several limitations affect practical implementation. The time required for detailed error analysis can be prohibitive in typical teaching contexts where instructors manage large classes and heavy workloads. Additionally, while error taxonomies provide excellent diagnostic tools, they may not fully capture the developmental nature of language acquisition or account for successful language production (Jabeen et al., 2015). The complexity of the classification system may also prove challenging for students to understand and apply in their own writing development.

Perhaps most significantly, the ways in which errors are presented and the manner in which they are discussed can profoundly affect students. Wang & Troia (2023) found that focusing too heavily on error analysis risks may create a deficit-oriented approach to writing instruction that may negatively impact student motivation and confidence. Therefore, while error analysis provides valuable insights for instruction, it should be balanced with approaches that recognize, emphasize, and build upon students' developing competencies in written expression.

2.4. Error Analysis and Artificial Intelligence

Recent developments in AI-powered writing tools have transformed approaches to error detection and analysis. As Godwin-Jones (2022) notes, modern AI systems can not only identify errors but also provide sophisticated correction suggestions and writing improvements. Tools like ChatGPT demonstrate capabilities far beyond traditional rule-based error detection, offering context-aware analysis of writing problems ranging from basic grammar to complex stylistic issues.

Current machine learning methods for error classification utilize Large Language Models (LLMs) that can identify and categorize multiple error types simultaneously. These systems can assess various aspects of writing including academic tone, cohesion, and discourse marker usage (Tate et al., 2023). The technology is particularly powerful for EFL/ESL contexts, where it can identify language-specific errors and provide targeted feedback. Some systems can even detect and correct errors while maintaining the writer's intended meaning, a significant advancement over previous rule-based approaches.

However, significant limitations and challenges persist in existing systems. As Krishna et al. (2023) demonstrate, AI tools can be inconsistent in their error detection and sometimes generate false positives. The accuracy of these systems varies considerably across different types of errors and writing contexts. Moreover, current research suggests that while these tools excel at identifying surface-level errors, they struggle with more nuanced aspects of writing assessment. Chiu et al. (2023) presented a systematic review of literature describing the challenges facing academia as AI tools are implemented, emphasizing the urgent need for further research to guide the appropriate use of these tools in educational settings, particularly regarding their reliability in error analysis and assessment. To avoid inadvertently creating more errors through the use of AI tools, Tseng & Warschauer (2023) propose a five-stage process for teaching students to use AI tools effectively: understand, access, prompt, corroborate, and incorporate.

3. Methodology

3.1. Taxonomy

An understanding of the terminology used in this paper to describe the taxonomy tree is essential before proceeding further (refer to Figure 1 for a detailed illustration). Tier 1 (T1) represents the top tier of the taxonomy, encompassing the primary error categories: Grammar Words (GW), Grammar Sentences (GS), and Spelling (SP). Each T1 category branches into subcategories, designated as Tier 2 (T2), which are further divided into specific error conditions at Tier 3 (T3).

For instance, the T1 category Grammar Words (GW) includes thirteen T2 subcategories (GW1 to GW13), covering demonstratives, adjectives, nouns, and other grammatical components. Each of these T2 subcategories is subdivided into precise T3 error conditions. GW1, Demonstratives, for example, contains five specific T3 error conditions (GW1A to GW1E), such as unclear references, ambiguous usage, and redundant application, each accompanied by definitions and illustrative examples.

A comprehensive review of the literature revealed no existing taxonomy with comparable depth or scope. Our initial framework drew from established linguistic theories, as previously discussed, and was subsequently expanded using Claude Sonnet 3.5 to generate subcategories, definitions, and examples. The current taxonomy organizes writing errors into three primary categories, but future enhancements will incorporate the analysis of style and discourse, expanding the taxonomy as advancements in LLM capabilities permit. Depending on experimental outcomes, additional primary categories may be introduced, or existing ones may be consolidated to optimize the system's efficiency.

>12A description of the organization of the current taxonomy is as follows: Spelling errors (SP) encompass mechanical errors in letter arrangement and formation unrelated to grammar, such as common misspellings and typographical errors. Grammar errors are subdivided into two major categories: word-level errors (GW) and sentence-level errors (GS). Word-level errors include eight subcategories: demonstratives, adjectives, nouns, prepositions, pronouns, word choice, articles, and irregular verbs. Each subcategory contains multiple detailed subclassifications - for instance, itemized pronoun errors include unclear antecedents, case errors, number agreement, and reflexive pronoun misuse. Sentence-level errors comprise the following subcategories: fragments, word order, verb forms and tenses, subject-verb agreement, capitalization, and compound-complex sentence structure. Sentence errors also include a punctuation category which addresses various mechanical errors including comma splices, apostrophe misuse, semicolon errors, and quote mark placement. Each category and subcategory is accompanied by clear definitions, specific error descriptions, and representative examples to ensure consistent application of the taxonomy during analysis.

A typical example (see Figure 1) is as follows: GW1 is the notation for grammatical errors identified with the word class "demonstratives," the pronouns this, that, these, and those. GW1A, the first error classification, identifies demonstrative errors involving "unclear references." The error code GW1A in the taxonomy includes that name ("unclear references"), along with a definition ("the demonstrative lacks a clear antecedent, making its referent ambiguous"), and an example which contains a sentence with the error and a brief explanation of why the sentence is incorrect ("The company invested in new software and revised their training program, but this frustrated the employees. → 'This' could refer to either the software investment or the training revision, making its antecedent unclear.").

The taxonomy attempts to provide a comprehensive framework for detailed error analysis. Each error type is accompanied by specific definitions, examples, and clear guidelines for classification to ensure consistent application across different texts and assessors. As shown in Figure 1 below, the taxonomy also includes a set of hierarchy rules and conflict resolution protocols to address cases where errors might fall into multiple categories. Often an error would return two or three different error types interchangeably. Fixing this issue was difficult, since some errors could be only one or the other but some could plausibly be identified as both. This led to the creation of the "hierarchy" of errors, an attempt to give supremacy to error types that are more significant grammatically. The process of refining the taxonomy categories and the hierarchy is ongoing and may take some time to complete.

Another issue involved spelling errors. It soon became obvious during testing that identifying and rectifying spelling errors prior to attempting grammatical analysis was critical. This was also true of certain types of punctuation, especially apostrophes, which effectively function as morphological elements in word types like possessives and contractions. These are just a few of the challenges we face in creating a complete and functional taxonomy of errors.

Figure 1. Example of taxonomy entry: GW, GW1, and GW1A-B. The taxonomy is written in JSON format to facilitate comprehension by the LLM(s).

```json
{
  "taxonomy": {
    "metadata": {
      "version": "Tax 3L_5.2 DS",
      "focus": "Comprehensive grammatical/spelling error taxonomy",
      "revision_date": "284-01-2025",
      "hierarchy_rules": [
        "1. Process spelling (GM) first",
        "2. Syntax errors (GS) override word-level errors (GW)",
        "3. Specific subcategories override parent codes",
        "4. Punctuation (GS6) supersedes possessives (GW3E)",
        "5. Modals (GW11) override tense errors (GS3)"
      ],
      "exclusion_criteria": {
        "spelling": "Excludes homophones handled by GW6 (Word Choice)",
        "punctuation": "Excludes typographical/stylistic punctuation",
        "verbs": "Modals (GW11) supersede tense errors (GS3)"
      }
    },
    "error categories": {
      "GW": {
        GW1": {
          "name": "Demonstratives",
          "description": "Errors involving demonstrative pronouns (e.g., 'this,' 'that,' 'these,' 'those').",
          "subcategories": {
            "GW1A": {"name": "Unclear references",
              "description": "The demonstrative lacks a clear antecedent, making its referent ambiguous.",
              "example": "The company invested in new software and revised their training program, but **this** frustrated the employees. → 'This' could refer to either the software investment or the training revision, making its antecedent unclear."},
            "GW1B": {"name": "Redundant Use",
              "description": "Unnecessary repetition of demonstratives with redundant terms (e.g., 'this here').",
              "example": "**This** book **here** is great. → 'Here' is redundant with 'this.'"                       }, [...]
```

153.2. Coding and APIs

The code written for this experiment implements a comprehensive document analysis system that uses an API call to an LLM (typically Claude Sonnet 3.5 or, in some cases, DeepSeek R1) for error detection and correction in text documents. At its core, the system is built around a Document Analysis System (DAS) module written in the Python coding language, which operates as the main workhorse, initializing with an API key (to access and prompt one of the above mentioned LLMs) and then using that to analyze a file containing error types using specific instructions.

The main processing workflow begins by reading an input text file, which is then split into manageable chunks by the DAS, typically one sentence per chunk. For each chunk, the DAS constructs a detailed prompt that combines the text to be analyzed with the cached comprehensive JSON taxonomy of error types. This taxonomy, spanning over 1100 lines, defines a hierarchical structure of error categories. The DAS then prompts the LLM to analyze the text using this taxonomy, identifying errors and assigning specific error codes from the taxonomy.

When the chunk is sent to the LLM, a prompt instructs it to apply the taxonomy's error codes precisely. The system creates this structured interaction through a two-part prompt: first, a system message is sent that establishes the LLM as a strict error correction tool, accompanied by the complete error taxonomy encoded in a codes_text variable. Second, a user prompt is sent that contains both the text to be analyzed and specific formatting instructions. The prompt requires the LLM to provide responses in a strictly defined format: the original text, followed by the corrected version, and then a numbered sequence of errors with their corresponding taxonomy codes and brief explanations. This structured format requires the LLM to list errors like "SP1A" for spelling issues or "GW1A" for demonstrative errors, ensuring consistent categorization. To maintain precision, the system sets the "temperature" parameter to 0.0, eliminating randomness in the LLM's responses ("Temperature" essentially controls the LLM's creativity, with 0 equalling deterministic responses and 1 equalling random responses.). The prompt also includes an example demonstrating the exact format expected for error reporting. After the response from the LLM is returned, the DAS carefully tracks any missing sentences for repeated analysis, and the results are written to an output file.

3.3. Pilot testing

Pilot testing (and the system's repeated failure) led to many refinements in both the taxonomy and the DAS (both the prompts and codebase). The following sections will detail the development timeline, key shifts in methodology, and experimental adjustments.

During initial testing, redundancy and conflicts between error types were frequent. Errors were often categorized under multiple classifications, complicating the analysis process. Addressing this required the development of the aforementioned error "hierarchy" to prioritize more grammatically significant error types. Refining this hierarchy remains an ongoing process.

Initial testing with a simplified Tier 2 (T2) taxonomy and a basic code framework demonstrated that LLMs could categorize writing errors effectively. To quantify accuracy, a separate program was developed to compare LLM output with predetermined error codes, highlighting exact matches, close T2 matches, and outliers. Encouraged by near-perfect initial results on T2 testing, the taxonomy was expanded to Tier 3 (T3), with sample sentences generated by ChatGPT 4.0 mini and Claude Sonnet 3.5. Early non-randomized T3 testing showed over 80% accuracy. However, randomizing the order of the test sentences resulted in a significant accuracy drop to around 60%, revealing issues in the taxonomy's robustness and the influence of human error during taxonomy refinement. The increase from 17 categories in the T2 taxonomy to 146 subcategories in the T3 taxonomy complicated manual curation of both error types (to reduce redundancies) and test sentences.

Despite these initial concerns that the taxonomy would need to be simplified to achieve consistent results, testing with a newly developed LLM, the DeepSeek R1 model, yielded much improved results (81-93% T2 match), demonstrating better reasoning capabilities compared to ChatGPT and Claude. These promising results prompted the development of a new T2/T3v5 taxonomy with DeepSeek's assistance, which incorporated additional subcategories, definitions, and examples. This expansion increased the sample size to approximately 860 sentences, complicating the curation process further. Before testing with DeepSeek using this new taxonomy and the expanded test sentence corpera could be tested, DeepSeek R1 was disrupted by denial-of-service attacks, necessitating a return to Claude, which produced mixed results (68-78% T3 matches). These outcomes underscored the need for better error sentence samples and refined code identifiers.

It is important to note that while LLMs frequently return correct error definitions, discrepancies often occur at the subcategory level. Errors are typically accurate at the broader T2 level even when T3 matches are inconsistent. Outliers, while divergent, often reveal overlapping error characteristics upon closer examination, highlighting the complexity of achieving precise, granular error classification.

3.4. Final testing

For the final phase of testing, we selected José da Fonseca's *English as she is spoke*, a peculiar 19th-century Portuguese-English phrasebook that has become infamous for its extraordinary number of linguistic errors. Originally published in

191855, the book was intended to help Portuguese speakers learn English. However, Fonseca, who reportedly spoke no English, created the book by using a Portuguese-French phrasebook and a French-English dictionary, resulting in a text riddled with systematic translation errors at multiple linguistic levels. The book was a sensation in English speaking countries for its comedic value, with Mark Twain penning the introduction for the American printing.

    The value of this text for our research lies in its unique characteristics. Unlike artificially created error sets or even genuine student writing samples, *English as she is spoke* provides a rich corpus of naturally occurring translation errors that span multiple linguistic categories. The errors in the text are both systematic and varied, appearing at morphological, syntactic, and semantic levels. This makes it an ideal test case for our error classification system, as it allows us to examine how well our taxonomy captures and categorizes different types of language transfer errors that occur through mechanical translation processes.

    We specifically selected Dialogue 16 from the text for our analysis because it contains a representative sample of the book's characteristic errors while remaining concise enough for detailed examination. The dialogue, which ostensibly attempts to model a conversation between a tourist and a guide, contains errors in nearly every sentence, ranging from basic grammatical mistakes to complete semantic breakdown. This density and variety of errors provides an

excellent test bed for evaluating our system's ability to identify and categorize multiple error types within single sentences.

Figure 2. Dialogue 16 (excerpt) from English as she is spoke (Fonseca, 2004)

| Pâra vêr â cidáde. | For to see the town. |
|---|---|
| António, acompânha êstes senhôres, ê móstra-lhes â cidáde. | Anthony, go to accompany they gentilsmen, do they see the town. |
| Desejâmos vêr ô quê élla contêm dê curiôso. | We won't to see all that is it remarquable here. |
| Tênhão â bondáde dê vír comígo; hêi-de mostrár-lhes quánto é merecedôr dê súa attenção. Êis-nos âo pé dâ cathedrál. Quérem entrár n'élla? | Come with me, if you please. I shall not folget nothing what can to merit your attention. Here we are near to cathedral; will you come in there? |
| Vêl-â-hêmos primeirâmênte pôr fóra ê depôis pôr dêntro. | We will first to see him in oudside, after we shall go in there for to look the interior. |

Unlike the contrived test sentences used in pilot testing that only test one error type per sentence, *English as she is spoke* presents authentic, albeit unintended, linguistic errors that span across multiple categories, from basic spelling and grammatical mistakes to complex sentence structure issues. It also contains sentences with multiple, often co-contributing errors, thus more accurately imitating the frequency and combinative nature of errors in student writings. The text is replete with errors ranging from incorrect infinitive constructions (e.g., "for to see the town") to subject-verb agreement issues (e.g., "it have ten archs"). This diversity allows us to thoroughly evaluate the system's error detection capabilities within the framework of our taxonomy. The system analyzed Dialogue 16 from the book, which is as follows:

> Title: For to see the town.
>
> A: Anthony, go to accompany they gentilsmen, do they see the town.
> B: We won't to see all that is it remarkable here.
> C: Come with me, if you please. I shall not folget nothing what can to merit your attention. Here we are near to cathedral; will you come in there? We will first to see him in outside, after we shall go in there for to look the interior. Admire this master piece gothic architecture's. The chasing of all they figures is astonishing indeed. The cupola and the nave are not less curious to see.
> B: What is this palace how I see youder?
> C: It is the town hall.
> B: And this tower here at this side?
> C: It is the Observatory. The bridge is very fine, it have ten archs, and is constructed of free stone. The streets are very layed out by line and too paved.
> B: What is the circuit of this town?
> C: Two leagues.
> B: There is it also hospitals here?
> C: It not fail them.
> B: What are then the edifices the worthest to have seen?
> C: It is the arsenal, the spectacle's hall, the Custom-house, and the Purse. We are going too see the others monuments such that the public pawn-broker's office, the plants garden's the money office's, the library.
> B: That it shall be for another day; we are tired.

## 4. Results

### 4.1. DAS Results

22Using Claude 3.5 Sonnet, the system successfully identified multiple layers of errors within each sentence. Figure 3 shows a sample of the returned analysis in its raw form.

Figure 3. Example Output: T3 Analysis of English as she is spoke Dialogue 16 (excerpt)

```
OT: For to see the town.
Corrected: To see the town.
1st Error & Reason: GW12A, for to see, Incorrect infinitive construction with redundant preposition
2nd Error & Reason: GS1D, sentence, Fragment lacking main clause or complete thought
OT: Anthony, go to accompany they gentilsmen, do they see the town.
Corrected: Anthony, accompany these gentlemen to show them the town.
1st Error & Reason: SP1A, gentilsmen, Spelling error in "gentlemen"
2nd Error & Reason: GW5A, they, Incorrect pronoun case - should be "these"
3rd Error & Reason: GS1D, do they see, Incomplete thought structure
4th Error & Reason: GS2A, go to accompany, Incorrect word order and redundant "go to"
OT: We won't to see all that is it remarkable here.
Corrected: We want to see all that is remarkable here.
1st Error & Reason: GW6A, won't, Confused word choice between 'won't' (will not) and 'want'
2nd Error & Reason: GS2A, is it, Incorrect word order with redundant 'it'
OT: Come with me, if you please.
Corrected: Come with me, if you please.
[No errors]
OT: I shall not folget nothing what can to merit your attention.
Corrected: I shall not forget anything that can merit your attention.
1st Error & Reason: SP1A, folget, Spelling error
2nd Error & Reason: GS2E, not...nothing, Double negative
3rd Error & Reason: GW6A, what, Incorrect relative pronoun
4th Error & Reason: GW6A, can to merit, Incorrect verb pattern with modal
OT: Here we are near to cathedral; will you come in there?
Corrected: Here we are near the cathedral; will you come in?
1st Error & Reason: GW7A, near to cathedral, Missing article before singular noun
2nd Error & Reason: GW4C, to, Redundant preposition with 'near'
3rd Error & Reason: GW1D, there, Unnecessary demonstrative reference
OT: We will first to see him in outside, after we shall go in there for to look the interior.
Corrected: We will first see him outside, then we will go inside to look at the interior.
1st Error & Reason: GW11A, will first to see, Incorrect modal construction with 'to'
2nd Error & Reason: GW4C, in outside, Redundant preposition
3rd Error & Reason: GW11A, shall go, Unnecessary formal modal 'shall'
4th Error & Reason: GW4B, look the interior, Missing preposition 'at'
5th Error & Reason: GW4C, in there, Redundant preposition
6th Error & Reason: GW4C, for to look, Redundant preposition 'for' with infinitive
OT: Admire this master piece gothic architecture's.
Corrected: Admire this masterpiece of Gothic architecture.
1st Error & Reason: GW3D, master piece, Incorrect compound noun formation - should be one word
2nd Error & Reason: GW2A, gothic architecture, Proper adjective 'Gothic' needs capitalization
3rd Error & Reason: GW3E, architecture's, Incorrect possessive form - no possession intended
4th Error & Reason: GW4B, masterpiece gothic, Missing preposition 'of' between nouns
```

4.2. Discussion

4.2.1. H1: AI Error Identification and Classification

The AI system's performance in identifying and classifying errors was assessed using eight sentences from *English as she is spoke* (Dialogue 16). The analysis of these sentences aimed to evaluate the system's ability to detect and categorize errors in a text with extensive linguistic flaws, using a predefined taxonomy.

Overall, the AI system identified a total of 32 errors across the 8 sample sentences analyzed above. Of these, 27 errors (84.4%) were correctly identified and classified according to the taxonomy, while three errors (9.4%) were partially correct in classification, meaning they were correctly flagged but assigned an incorrect subcategory. Two errors (6.2%) were either misclassified or entirely overlooked. One sentence was correctly determined to be mistake free.

In the sentence "For to see the town," the AI correctly identified the erroneous infinitive construction, labeling it as an incorrect structure. However, it also flagged the sentence as a fragment, which was unnecessary given that it functioned as a title. In "Anthony, go to accompany they gentilsmen, do they see the town," the AI successfully detected multiple errors, including a spelling mistake, an incorrect pronoun case, and improper word order. However, it misclassified the phrase "do they see the town" as an incomplete thought, when in fact, the issue was related to structure and syntax rather than being a fragment.

Another instance of effective AI classification occurred in the sentence "We won't to see all that is it remarkable here."The system correctly flagged the incorrect word choice and the improper word order. However, it failed to detect the redundancy of "all that is it remarkable." In contrast, the AI correctly determined that "Come with me, if you please." contained no grammatical errors, though it did not account for the phrase's archaic and non-standard usage.

A more complex case was observed in "I shall not folget nothing what can to merit your attention." Here, the AI effectively identified and corrected a spelling error, a double negative, and an incorrect relative pronoun. However, it mistakenly classified the relative pronoun issue as a word choice error rather than a grammatical misclassification. Additionally, the AI miscategorized the phrase "can to merit" under an invented new more accurate *modal with infinitive* subcategory rather than recognizing it as a standard verb pattern error. Classifying this as an error by the AI system may be a bit unfair, in fact.

The AI also demonstrated success in correcting "Here we are near to Cathedral; will you come in there?" by properly identifying a missing article and an unnecessary preposition. However, it misclassified the redundant demonstrative "there."

One of the most error-dense sentences was "We will first to see him in outside, after we shall go in there for to look the interior." The AI correctly flagged six distinct errors: the incorrect modal construction, two redundant prepositions, an unnecessarily formal modal, a missing preposition, and an incorrect pronoun reference. While the AI performed well in categorizing these errors, it failed to recognize the larger issue of coherence across clauses.

Finally, in "Admire this master piece gothic architecture's," the AI effectively identified errors in noun formation, capitalization, possessive structure, and missing prepositions. However, it misclassified the capitalization issue under an incorrect subcategory.

In summary, the DAS demonstrated a high level of accuracy in detecting and categorizing linguistic errors, though it occasionally misclassified subcategories and overlooked contextual coherence. While the system successfully corrected most structural and lexical mistakes, its limitations in handling redundancy, referential ambiguity, and discourse-level cohesion suggest areas for future refinement. The overall findings indicate that AI-assisted error classification is a promising tool for EFL writing instruction, with the potential for further improvement through refined error taxonomies and enhanced contextual processing.

4.2.2 Hypothesis 2: Expert Confirmation of AI Error Classification

To evaluate the accuracy of the AI's error identification and classification, a panel of three expert linguists (PhDs in Applied and Theoretical Linguistics) independently reviewed the AI's analyses. Each expert was provided with the original sentences, the AI-generated error classifications, and the corrected sentences. The panelists were tasked with verifying whether the AI's classifications were correct, partially correct, or incorrect based on the predefined taxonomy. Results of the expert analysis are as follows:

1. For to see the town. Correction: To see the town. The system correctly identified the incorrect infinitive construction (GW12A) "for to see" and flagged the sentence as a fragment (GS1D) lacking a complete thought. Had it correctly identified this as a title it may have more accurately forgiven the fragment. This points to an unfortunate side effect of "chunking," feeding the LLM one sentence at a time. It may not see the role a given piece of the text plays in the larger context.
2. Anthony, go to accompany they gentilsmen, do they see the town. Correction: Anthony, accompany these gentlemen to see the town. The AI flagged four errors: a spelling mistake (SP1A) in "gentilsmen," incorrect pronoun case (GW5A) with "they," incomplete thought structure (a "fragment error" GS1D) in "do they see," and redundant phrasing (GS2A) in "go to accompany." Of these errors, the fragment error is suspect. It should more accurately be identified as an infinitive error in view of the corrected version.

3. We won't to see all that is it remarkable here. Correction: We want to see all that is remarkable here. The system identified the confused word choice (GW6A) of "won't" instead of "want" and the incorrect word order (GS2A) in "is it remarkable." It failed to flag a possible redundancy created by "all that is it remarkable," where "it" is unnecessary, possibly because an error had already been flagged in that phrase.

4. Come with me, if you please. No errors were detected, and the sentence was correctly marked as grammatically sound, even though the phrase "if you please" is an archaic and nonstandard phrasing of this type of politeness.

5. I shall not folget nothing what can to merit your attention. Correction: I shall not forget anything that can merit your attention. The AI identified four errors: the spelling error (SP1A) in "folget," the double negative (GS2E) "not...nothing," the incorrect relative pronoun (GW6A) "what," and the improper verb pattern (GW6A) "can to merit." The analysis was thorough, and the system wasn't confused by the archaic use of the modal "shall not" for "won't." However, the codes for the final two errors were themselves incorrect. "Incorrect relative pronoun" should be coded as GW5E. The error marked as "improper verb pattern (GW6A)" is very interesting. It could possibly have been identified as some type of "error in using modal verbs (GW11). " However, no subtype in that category matches this type of hybrid infinitive construction. Seeking a match it invented "incorrect verb pattern with modal," a category that doesn't even exist, but which is similar to G12A, "incorrect verb patterns," under W12 "Gerunds and infinitives," which describes using a gerund where an infinitive is called for or vice versa. This type of "category invention" occurs again later with the same modal/infinitive error, but with different results (see sentence 7 below).

6. Here we are near to cathedral; will you come in there? Correction: Here we are near the cathedral; will you come in? The system flagged the missing article (GW7A) "the" before "cathedral," the redundant preposition (GW4C) "to," and the unnecessary demonstrative (GW1D) "there." The awkward phrasing "will you come in there" could probably also have been flagged. The code for the "unnecessary demonstrative" is incorrect; however, no subcategory among demonstrative errors accurately matched this error, so it probably became confused. GW1D is actually the code for "distance confusion." "Redundant use of demonstratives (GW1B)" or "overuse of demonstratives (GW1E)" probably would have been closer matches. This points to the ways in which "holes" in the taxonomy can be discovered and new error categories can be coded through exposure to more textual errors in future research.

7. We will first to see him in outside, after we shall go in there for to look the interior. Correction: We will first see him outside, them we will go inside to look at the interior. The AI identified six errors: incorrect modal construction with "to" (which it codes as GW11A) "will first to see," redundant preposition (GW4C) "in outside," unnecessary formal modal (which it also codes as GW11A) "shall," missing preposition (GW4B) "at" in "look the interior," and redundant prepositions (2xGW4C) "in there" and "for to look." These corrections were very interesting, because it actually renamed a category (GW11A) TWICE to account for error types it couldn't find. First it renames it "incorrect modal construction with to," a category that doesn't exist. And although it had previously allowed the archaic "shall" it now flagged it as "overly formal" (interestingly, again renaming the category GW11A, which is actually "tense sequence error"). It is worthwhile to note here that due to the "chunking" of the text (feeding each sentence to the LLM out of context) it would not be "aware" of this earlier leniency with "shall." More importantly, this sentence represents the first failure in the correction phase. The system failed to identify the gendered "him" assigned to the noun "cathedral." Perhaps because the antecedent was in a previous sentence the system assumed that this was referring to a person. This points to the need to incorporate a contextual awareness into the system in the future.

8. Admire this master piece gothic architecture's. Correction: Admire this masterpiece of Gothic architecture. The system correctly flagged the compound noun error (GW3D) "master piece," correcting it to "masterpiece," the capitalization (which it miscodes as GW2A, perhaps because there is no subcategory of GS5 "Capitalization" that deals with proper nouns, specifically) of the proper noun "Gothic," the unnecessary possessive apostrophe (GW3E) in "architecture's," and the missing preposition (GW4B) "of."

Despite the promising results, some limitations were noted. The system occasionally struggled with overlapping error categories, particularly when errors could be classified under multiple subcategories. For example, redundancy and incorrect word choice sometimes conflicted, leading to inconsistent classifications. Additionally, while the system performed well with isolated sentences, it showed less accuracy when dealing with contextual nuances across multiple sentences, an area identified for future refinement.

Perhaps most interesting of all were the findings that could be inferred from the coding errors. When the system was unable to find a match in the taxonomy for an error it did one of two things: it either tried to find the closest match (an effort that often caused it to miscode a correct error description), or it invented a new error code entirely and used an existing code to flag it. This second case is fascinating, since it appears that the LLM is not only able to correct the taxonomy, but that when given no other option it actually attempts to do so.

The use of *English as she is spoke* for final testing provided valuable insights into the strengths and weaknesses of our AI-assisted error analysis system. The results highlight the system's ability to handle complex, multi-layered errors while also pointing to areas where further development is needed to enhance contextual understanding and error classification consistency.

The results of the expert verification process demonstrated full agreement with the analysis of the AI results from the eight sentences above. All three linguists independently confirmed that the DAS identified every error in the text (32), that the DAS was accurate in identifying 27 of these errors, that it was partially correct in identifying three of the errors, and that two errors were overlooked or misclassified.

This verification of the results by human experts strongly supports H2. The consensus among the expert panel indicates that the AI was capable of applying the error taxonomy with moderate precision and reliability. Additionally, this validation underscores the potential for AI-driven error analysis tools to serve as robust supplementary resources for EFL instruction, providing consistent and replicable error identification without reliance on subjective human interpretation.

While these results affirm the system's effectiveness, further improvements in both the taxonomy and codebase are needed, and future research is needed to explore whether similar accuracy rates hold across more diverse linguistic samples, including student-generated texts with varying proficiency levels. Additionally, future iterations of the AI system may incorporate contextual refinements to address its occasional limitations in discourse-level analysis.

Nonetheless, the findings show that AI-assisted linguistic error analysis may someday achieve expert-level accuracy, positioning it as a viable tool for individuated writing feedback and assessment in educational settings.

5. Conclusion

5.1. Summary of findings

This study demonstrated the potential of AI-assisted error analysis in identifying and categorizing linguistic errors in EFL writing. Using a detailed taxonomy, the AI system effectively classified errors at the word and sentence levels, providing granular feedback beyond traditional holistic assessments. The final testing phase, conducted using *English as she is spoke*, confirmed the AI's ability to detect diverse error types, including grammatical, lexical, and structural mistakes. However, limitations were observed in contextual understanding and consistency in error classification, particularly when errors overlapped multiple categories. Notably, the AI occasionally generated new error classifications when encountering previously undefined issues, indicating both the strengths and limitations of current AI-driven linguistic analysis. These findings highlight the potential for AI to enhance EFL writing instruction while underscoring the need for continued refinement of error taxonomies and AI prompting strategies.

5.2. Practical applications

The error classification system developed in this study offers several immediate applications for EFL writing instruction. Teachers can use this AI-assisted approach to provide more detailed and consistent feedback on student writing while significantly reducing the time required for error analysis. The system's ability to identify and categorize specific error types enables instructors to develop targeted intervention strategies and create materials that address common error patterns in their classes.

The hierarchical nature of the taxonomy allows for flexible implementation based on student level and instructional needs. At lower levels, teachers might focus on mechanical and word-level errors, while advanced writing instruction can incorporate more complex sentence-level analysis. The system can also help instructors track student progress over time by monitoring changes in error patterns, enabling more informed decisions about curriculum development and instructional focus.

5.3. Directions for future research

This study suggests several promising avenues for future research in AI-assisted error analysis. The most immediate need is to apply this system to actual student writings across various proficiency levels and linguistic backgrounds. Such research would help validate the taxonomy's effectiveness in real educational contexts while potentially revealing necessary adjustments to the classification system.

Further research should also explore expanding the system's analytical capabilities beyond grammatical errors to include assessments of style, tone, and cultural fluency. This expansion would address the growing recognition that effective writing instruction must consider both mechanical accuracy and communicative effectiveness. The development of reliable AI-based methods for analyzing these more nuanced aspects of writing would represent a significant advancement in automated writing assessment.

The addition of multiple descriptive dimensions to error analysis also warrants exploration. Future research could incorporate classifications based on error severity, error source (following Corder's framework), and impact on communication. These additional analytical layers would provide a more nuanced understanding of student writing development and help prioritize areas for intervention.

A particularly crucial area for future research involves the development of training data through extensive corpus analysis. Large-scale studies of student writing errors would help identify gaps in the current taxonomy and refine error type descriptions. This work would be essential for improving the system's accuracy and comprehensiveness in error identification and classification.

Perhaps most importantly, research is needed to determine optimal methods for presenting error feedback to students. Future studies should investigate questions of feedback quantity and specificity, seeking to understand how much detailed error analysis students can effectively process and apply to their writing development. This research should particularly focus on making error feedback actionable, moving beyond simple correction to support genuine improvement in writing ability. The goal should be to develop feedback systems that enhance student awareness of error patterns and promote independent error identification and correction skills.

5.4. Limitations

This study encountered several significant limitations that should inform future research directions. These constraints manifested across multiple domains: technological capabilities, training data adequacy, testing scope, and human verification processes.

The technological limitations primarily stemmed from current constraints in Large Language Model capabilities. While contemporary LLMs demonstrate impressive analytical abilities, they face inherent restrictions in processing capacity and training data integration. Hardware constraints, particularly in locally-hosted implementations, further restricted the scope of analysis. Additionally, the current cost structure of API-based processing imposed practical limitations on the scale of implementation, though these constraints may diminish as the technology evolves.

The training data utilized in this study, while sufficient for initial validation, represents only a preliminary exploration of the potential training corpus. It did not test every error in the taxonomy. A more comprehensive training dataset would need to encompass a broader range of error types, more detailed descriptors, and an expanded set of examples. These enhancements would be crucial for developing a more robust and nuanced error classification system.

Testing limitations also affected the study's scope and generalizability. The analysis focused on a single short text, primarily due to the aforementioned technological constraints. This restricted testing scope limited our ability to refine the taxonomy through the identification and classification of novel error types. Furthermore, the system requires validation across a broader spectrum of texts, including student writings at various proficiency levels, different compositional modes, and diverse subject areas.

The human verification process presented additional limitations. The current study employed a relatively narrow verification approach, focusing primarily on the AI system's ability to identify and classify errors accurately. Future research would benefit from expanded human participation, particularly as the system's capabilities extend into more subjective areas of composition analysis, such as discourse patterns and stylistic elements. The limited scope of human verification, while appropriate for this initial study, suggests the need for more comprehensive validation procedures in future implementations.

41Richards, J. C. (1971). A non-contrastive approach to error analysis. *ELT Journal, XXV*(3), 204–219. https://doi.org/10.1093/elt/XXV.3.204

Richards, J. C. (Ed.). (1974). *Error analysis: Perspectives on second language acquisition*. Longman.

Seddik, M. (2023). Error analysis in EFL students' writing skill. *Journal of English Language Teaching and Applied Linguistics, 5*(4), 163–172. https://doi.org/10.32996/jeltal.2023.5.4.16

Shi, H., & Aryadoust, V. (2024). A systematic review of AI-based automated written feedback research. *ReCALL, 36*(2), 187–209. https://doi.org/10.1017/S0958344023000265

Tate, T. P., Doroudi, S., Ritchie, D., Xu, Y., & Uci, M. W. (2023). Educational research and AI-generated writing: Confronting the coming tsunami [Preprint]. *EdArXiv, January, 10*. https://doi.org/10.35542/osf.io/4mec3

Tseng, W., & Warschauer, M. (2023a). AI-writing tools in education: If you can't beat them, join them. *Journal of China Computer-Assisted Language Learning, 3*(2), 258–262. https://doi.org/10.1515/jccall-2023-0008

Wang, H., & Troia, G. A. (2023). How students' writing motivation, teachers' personal and professional attributes, and writing instruction impact student writing achievement: A two-level hierarchical linear modeling study. *Frontiers in Psychology, 14*, 1213929. https://doi.org/10.3389/fpsyg.2023.1213929